\documentclass[conference]{IEEEtran}
\ifCLASSINFOpdf
  \usepackage[pdftex]{graphicx}
  \graphicspath{{images/}}
  \DeclareGraphicsExtensions{.pdf,.jpeg,.png,.jpg}
\else
\fi
\usepackage{amsmath}
\usepackage{algorithm, algpseudocode}

\usepackage{siunitx}

\begin{document}
%
\title{Hybrid Forests for Left Ventricle  Segmentation using only the first slice label}




%
\author{
    \IEEEauthorblockN{Isma\"el Kon\'e and
                      Lahsen Boulmane\IEEEauthorrefmark{1}
                      }
    \IEEEauthorblockA{2MIA Research Group , LEM2A Lab,\\ Faculty of Sciences, Moulay
                        Ismail University,\\BP 11201, Avenue Zitoune\\ Meknes, Morocco
                     }
}

\makeatletter
\def\ps@IEEEtitlepagestyle{
  \def\@oddfoot{\mycopyrightnotice}
  \def\@evenfoot{}
}
\def\mycopyrightnotice{
  {\footnotesize
  \begin{minipage}{\textwidth}
  \centering
  \IEEEauthorrefmark{1} This article has been accepted at the 3rd International Conference \\on
                        Intelligent System and Computer Vision (ISCV 2018), Fez, Morocco.\\ 
                         978-1-5386-4396-9/18/\$31.00 \textcopyright2018 IEEE.
  \end{minipage}
  }
}


\maketitle

\begin{abstract} Machine learning models produce state-of-the-art results in many MRI images segmentation. However, most of these models are trained on very large datasets which come from experts manual labeling. This labeling process is very time consuming and costs experts work. Therefore finding a way to reduce this cost is on high demand. In this paper, we propose a segmentation method which exploits MRI images sequential structure to nearly drop out this labeling task. Only the first slice needs to be manually labeled to train the model which then infers the next slice's segmentation. Inference result is another datum used to train the model again. The updated model then infers the third slice and the same process is carried out until the last slice. The proposed model is an combination of two Random Forest algorithms: the classical one and a recent one namely Mondrian Forests. We applied our method on human left ventricle segmentation and results are very promising. This method can also be used to generate labels.

\end{abstract}


%
\IEEEpeerreviewmaketitle

\section{Introduction}
In clinical practice, doctors use cardiac cine-MRI  for early detection of some cardiac attacks like heart failure with or without infarction. They manually delineate the Left Ventricle (LV) in each slice of the cine-MRI and use that for computing the Ejection Fraction percentage. Based on this metric, they diagnose the patient as presenting symptoms of a particular heart attack \cite{hf_assess}.
The bottleneck of this procedure is the manual delineation or labeling which is very time consuming.
Therefore, automation is the way to face this problem. Recently, the field of Machine Learning, especially Deep Learning (DL) has shown state-of-the-art in this kind of task\cite{dl_medical}. However it comes with a big price: the availability of huge amount of hand-labeled data for training the model.
Thus we have two challenges:
\begin{itemize}
  \item Automating the task of delineation.
  \item Reducing the burden of manual delineation to train a machine learning model.
\end{itemize}
To illustrate the first challenge, consider a doctor that receives daily 10 patients to diagnose a heart failure. We suppose each patient cardiac cine-MRI produces 10 slices, then the doctor has 10 x 10 = 100 slices to delineate on a daily basis. So this is a great part of the doctor's time that's lost on a redundant and tiring work. Therefore automation will be very helpful and allow the doctor to care more cases daily.

For the second challenge, following the first example,  suppose we decide to use deep learning model automate the task. Now we have to prepare a training dataset of cardiac cine-MRI, say 100 patients' cardiac cine-MRI  where each  patient cardiac cine-MRI produces 10 slices as before. So we must delineate the left ventricle on 100 patients x 10 slices, a total of 1000 slices to be able to train the deep learning model.  This is obviously a very huge and costly work in terms of resources and time.

Here we propose a simple framework that exploits the sequential nature of MR images to reduce this labeling burden. We use this framework with a combination of classical machine learning algorithms, namely Standard Random Forests (RF)\cite{randomforest} and Mondrian Forests (MF)\cite{mondrianforest} and results are very promising.
Our contribution are two folds:
\begin{itemize}
  \item presenting a simple framework to MRI images segmentation using few labels (Sect.~\ref{intuition}).
  \item show its effectiveness in the case of LV segmentation.
\end{itemize}
We apply this method on the 2017 Automated Cardiac Diagnosis Challenge (ACDC) dataset \cite{acdc} (Sect.~\ref{exp_res}). 
%
%

\begin{figure}[!t]
\centering
\includegraphics[width=2.5in]{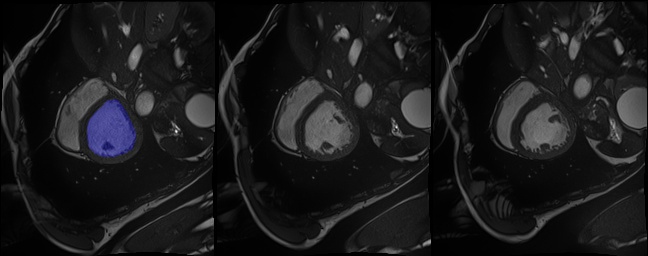}
\caption{Three consecutive slices of cine-MRI from ACDC challenge dataset. In the first slice, the Left Ventricle (LV) is overlaid in blue.}
\label{fig_3slic}
\end{figure}

\section{Dataset}
The dataset used in this study comes from the Automated Cardiac Diagnosis Challenge (ACDC) associated with the previous Medical Image Computing and Computer-Assisted Intervention (MICCAI) 2017\cite{acdc}.  
Data are Cine-MR images of the cardiac cycle of patients having some pathologies. They were fully anonymized and handled within the regulations set by the local ethical committee of the Hospital of Dijon (France). Labels in this dataset are segmentation of the heart in Right and Left Ventricles, Myocardium and the background. However in our experiments, we considered only the Left Ventricle segmentation which is mostly use for evaluating heat failure attacks\cite{hf_assess}. \figurename ~\ref{fig_3slic} shows a sample of 3 slices from a single cardiac cine-MR of the dataset. The overlaid region in the first slice is the Left Ventricle   (LV).

\section{Method}

\subsection{Intuition}
\label{intuition}
MRI images are a sequence of closed image sections of a part of the body that contains many organs. Consider looking at the image (\figurename ~\ref{fig_3slic}), we notice strong resemblance between the 3 successive slices.
When we observe the first slice and then we look at the second, we directly make associations between both slices.  Specifically, an expert may tell us for instance, the blue overlay area in the first slice is the left ventricle (LV) (\figurename ~\ref{fig_3slic}). Then if we look at the next  slice, we immediately notice an area in the second slice that has approximately the same shape, appearance and position within the slice than the first slice's LV. Therefore this area in the second slice is probably the LV. At the second slice, we update or refine our knowledge of how the LV looks like and use that to infer the third slice and the same process is carried out until the last slice.  This is done based on visual features: LV's position in the slice, its shape and gray intensities distribution over its surface which may present textures.

We can re-interpret this natural process in two steps:
\begin{itemize}
  \item A learning or training phase which is the observation of the segmented LV  from the first slice : the blue overlay (\figurename ~\ref{fig_3slic}). 
  \item An inference phase where we segment the second slice i.e   we find the part which is likely to be the LV.
\end{itemize}

So our strategy consists in building a machine learning algorithm that follows this simple scheme in order to segment the LV from cardiac cine-MRI. Practically, the algorithm will loop over these two steps at each slice. This approach is based on our previous work \cite{mriseg}.

\subsection{Formulation}

Given ${\{ S_{k}  /  k = 1, ... , N \}}$ representing the sequence of $N$ slices of a cardiac cine-MRI and considering a slice as a set of pixels, we aim at finding the subset of pixels namely $LV_{k}$  in each slice $S_{k}$ that belongs to the LV. And we denote by $Bg_{k}$ (Background) the subset of pixels in slice $S_{k}$ that is outside $LV_{k}$. So we formalize the problem as a binary pixel-wise classification where we assign a label $y$ to each pixel $x$ such that:

\begin{equation}
\label{y}
  y=\begin{cases}
    1, & \text{if $x \in$ LV (Left Ventricle)}.\\
    0, & \text{otherwise}.
  \end{cases}
\end{equation}

We're also given labels of the first slice.


\begin{algorithm}
  
  \begin{algorithmic}[1]
    \caption{SegmentLeftVentricle($\{ S_{k}/k = 1, ... , N \}, LV_{1} $)}
    \label{seglv}
      \State $\theta_{MF}\gets$ Learn $LV_{1}$ and $Bg_{1} = S_{1}\setminus LV_{1}$ using a ($MF$) Classifier.
      \State $\theta_{RF}\gets$ Learn $LV_{1}$ and $Bg_{1} = S_{1}\setminus LV_{1}$ using a ($RF$) Classifier.
      \For{$k = 1, ..., N-1$}
        \State $LV_{k+1}^{MF}  \gets \{x \in S_{k+1}/ Prob(x \in LV_{k+1} \mid \theta_{MF}) \geq   Prob(x \in Bg_{k+1} \mid \theta_{MF} )\}$.
        \State $LV_{k+1}^{RF}  \gets \{x \in S_{k+1}/ Prob(x \in LV_{k+1} \mid \theta_{RF}) \geq   Prob(x \in Bg_{k+1} \mid \theta_{RF} )\}$.
        \State $LV_{k+1}^{MF} \gets PostProcess(LV_{k+1}^{MF}$)
        \State $LV_{k+1}^{RF} \gets PostProcess(LV_{k+1}^{RF}$) 
        \State $LV_{k+1} \gets Combine(LV_{k+1}^{MF}, LV_{k+1}^{RF}) $
        \State $\theta_{RF}\gets$ Learn $LV_{k+1}^{MF}$ and $Bg_{k+1} = S_{k+1}\setminus LV_{k+1}^{MF}$  
      \EndFor\label{euclidendwhile}
      \State \textbf{return} $\{ LV_k/k = 1, ... , N \}$
  \end{algorithmic}
\end{algorithm}

We use algorithm \ref{seglv} to segment the LV. Inputs are cardiac cine-MRI slices and its first slice manually segmented or labeled by an expert.
We start by learning the left ventricle $LV_1$  and the background $Bg_1$ from the first slice . We do it with two different machine learning models: MF and RF Classifiers\cite{mondrianforest,randomforest}. The parameters $\theta_{MF}$ and $\theta_{RF}$ represent knowledge acquired from learning respectively with MF and RF. From that point we loop over next slices to infer the LV part. For each slice, we infer the subset of pixels that are part of the LV  with each model independently using the same criterion: the probability ($Prob$) of the pixel being in the LV of the current slice with the associated model knowledge must be greater than its probability of being in the background ($Bg$). Then we apply some post-process operations on the segmentation results of each model in oder to correct some inference errors. The LV of the current slice is the combination of  that has been inferred by each model. Now comes the update step which is one of the trickiest part. We use the inference or segmentation result of the MF Classifier ($LV_{k+1}^{MF}$) as a new data to train a new RF Classifier ($\theta_{RF}$) and we let the MF as is. This trick is what gives us better accuracy among possible update scenarios. Finally the LV of the cardiac cine-MR is the set of slice-wise segmented LV. 

Another important component in training a model is the choice of the features aka feature engineering. Following the previous part, we stated that our ability to recognize the LV in the next slice is based on the same shape, appearance and position within the previous slice. So here we choose the simple ones which are: 
\begin{itemize}
    \item the pixel value (visual appearance)
    \item the two coordinates  $(x, y)$  of a pixel (location)
\end{itemize}

\subsection{Post-processing}
\begin{figure}[!t]
\centering
\includegraphics[width=3.5in]{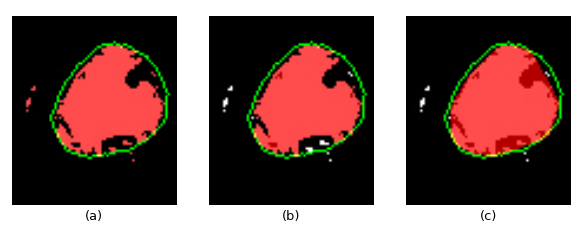}
\caption{MF and post-processing results of a slice. Green contour is the ground truth area delimiter.  (a) Red areas represent inference result of a MF. (b) Red area is the LV obtained after using the contour of maximum overlapping with previous slice. White blobs are areas discarded by this operation. (c) Red area is the resulting LV  after applying the convex hull on the LV found in (b). }
\label{postprocess}
\end{figure}

In \figurename ~\ref{postprocess}(a), red regions represent a typical result of a MF classifier inference.  The green contour delimits the ground truth area. Comparing both, we notice misclassified pixels outside (isolated blobs), inside (holes) and at the boundaries of the ground truth LV.
We propose two operations to resolve these issues:

\begin{itemize}
  \item Finding the contour whose area maximally overlaps with previous slice LV. It allows to eliminate isolated blobs and holes.
  \item Computing the convex hull of the result to recover missing pixels at the boundaries.
\end{itemize}

\subsubsection{Finding the area of maximum overlapping with previous slice LV}
the objective is to reclassify isolated blobs as background pixels and holes as LV pixels.
For that, we apply an edge detection technique \cite{findcontours} to retrieve all contours of the segmentation result. Then we select the contour whose area has the largest overlap with the previous slice's LV. This comes from the alignment property of MR images so the previous slice LV has the nearly the same location than the current slice one. Doing so discard isolated blobs' contour and reconsider holes in the final result. \figurename ~\ref{postprocess}(b) illustrates the process. We notice small holes are now red thus considered as the LV pixels and isolated blobs are white so discarded. 

\subsubsection{Convex hull}
After the previous operation, we still miss many pixels at boundaries. These missing pixels form concavities on the LV (see \figurename ~\ref{postprocess}(b)). The objective is to close these concavities to get even closer to the ground truth and thus reduce errors. As the LV ground truth is approximatively convex, we use the convex hull to close concavities on the contour. 
In order to compute the convex hull, the contour is considered as a polygon where each contour point is a vertex and each segment joining two consecutive vertexes is a side. Briefly, the convex hull is determined by computing the angle formed by a vertex and adjacent sides. If the angle is less than $180\si{\degree}$ or $\pi$ then we have a concavity so we suppress this vertex from the polygon. We do it using the method in \cite{convexhull}. \figurename ~\ref{postprocess}(c) shows the final result. We notice the red area includes the concavities.

\section{Experiments and results}
\label{exp_res}
We used a Intel i7 CPU core laptop. We run experiments on Ubuntu 16.04 environment with python 2.7 language programming. We used RF implementation from sklearn library\cite{sklearn} and OpenCV library\cite{opencv} python binding for post-processing operations. For the MF and RF classifiers, we set the number of estimators or trees to 50 and the minimum sample per leaf to 2.

We ran experiments in a couple of scenarios, each one starts by training each model i.e MF and RF with the first slice, then inferring subsequent slices following specific schemes:
\begin{itemize}
  \item Experiment 1: basic inference.
  \item Experiment 2: with post-processing.
  \item Experiment 3: with post-processing and updating as shown in algorithm \ref{seglv}. 
\end{itemize}
We randomly selected patient 11 of the ACDC challenge dataset and use the ED phase cine-MR to perform these experiments. The objective is to zoom in to analyze the method at the slice level. Then, we selected a sample of 45 patients and ran the same experiments.

\subsection{Qualitative results}
\begin{figure}[!t]
\centering
\includegraphics[width=3.5in]{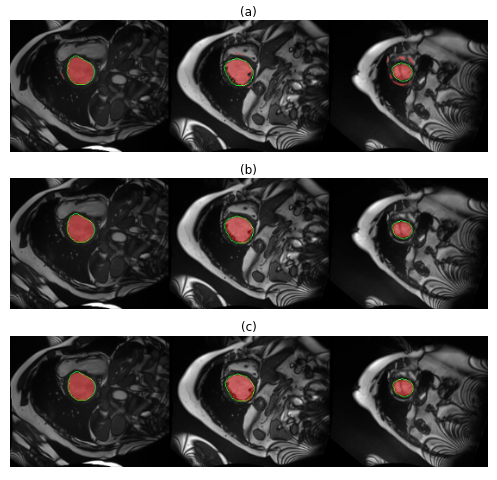}
\caption{Image results of the three experiments. Each image is composed of slices 2, 5, 8 of the ED phase cine-MR from patient 11 of the dataset. Green contour delimits the LV ground truth. Red overlay is the segmented LV found. (a) experiment 1: without post-processing and updating. (b) Experiment 2: with processing. (c) Experiment 3: with post-processing and updating.}
\label{res}
\end{figure}

Here, we present the visual results using a single patient (patient 11 of the ACDC challenge dataset) cardiac cine-MR.
In \figurename ~\ref{res}, we present results of each experiment previously stated. We present only three slices to avoid overloading the page, the 2nd, 5th and 8th slice. The green contour delimits the LV ground truth, meaning everything inside it is part of the LV. The red overlay is the segmented LV of our approach with respect to the experiment settings.

\subsubsection{\figurename ~\ref{res}(a)}
shows the result of experiment 1 where we don't process model inferences nor updating them. We notice the first slice in the figure is very good but the next is less good. In the second slice, holes inside the LV and some of its boundary parts are excluded from the segmentation result. Even worse, the last slice has some green overlay parts around the LV thus the algorithm is really misclassified these pixels as part of the LV.
 
\subsubsection{\figurename ~\ref{res}(b)} 
on its turn present experiment 2 results. We clearly see the impact of the post-processing step. Now all slices are nearly equally good, the ground truth and segmented LV are very closed.
Holes and boundaries are included in the LV conversely to experiment 1. Also we do not see any surrounding green overlay so they've been discarded.

\subsubsection{\figurename ~\ref{res}(c)} 
which presents the last experiment result doesn't show a noticeable difference from experiment 2. But in fact it adds a small amount on the accuracy. It's more evident in the next section where we use an evaluation metric.

\subsection{Quantitative results}

\begin{figure}[!t]
\centering
\includegraphics[width=3.5in]{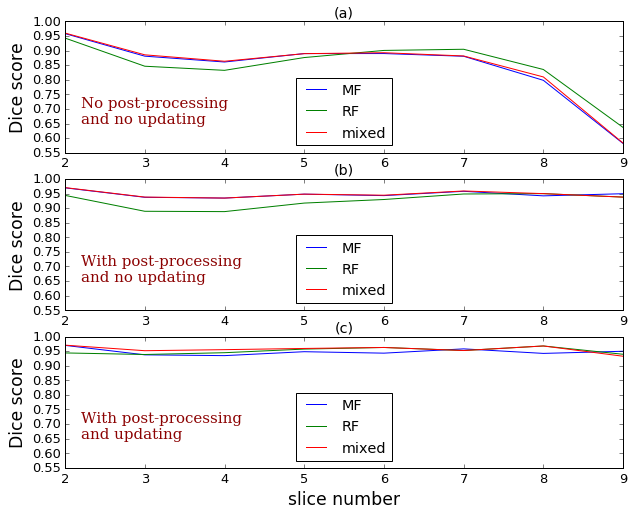}
\caption{Dice metric applied to each slice of the same cine-MRI than \figurename \ref{res} in the setting of experiment 1: without post-processing and update. The blue, green and red curves are respectively variation of dice score over slices for the MF, RF, and the combination of  both.}
\label{dice_g}
\end{figure}

\begin{table}[t]
\caption{Dice score mean and variance of three experiments on a sample of ED phase cine-MR of 45 patients from the dataset.}
\label{res_tab}
\begin{center}
\begin{tabular}{ |p{1.8cm}|p{1.8cm}|p{1.8cm}|p{1.8cm}| }
 \hline
 Experiments & Mondrian Forests & Random Forests & Combination of both\\
 \hline \hline
 Basic inference  &  $0.807\pm 0.063$  & $0.821\pm 0.065$ & $0.822\pm 0.065$ \\\hline
 With post-processing   & $\boldsymbol{0.9\pm 0.076}$  & $0.884\pm 0.074$ & $\boldsymbol{0.9\pm 0.077}$ \\\hline
 With updating   & \textendash & $\boldsymbol{0.901\pm 0.076}$  & $\boldsymbol{0.902\pm 0.077}$ \\
 \hline
\end{tabular}
\end{center}
\end{table}

To evaluate our results, we use the dice coefficient metric of two sets\cite{dice}:
\begin{equation}
\label{dice}
  Dice(A, B) = \frac{2 \lvert A \cap B\rvert}{\lvert A\rvert + \lvert B\rvert},
\end{equation}

where $\lvert A \rvert$ means the cardinal of the set $A$.
It's a similarity measure between two sets and its varies between 0 and 1 where 0 means the two sets are totally different and 1 means a perfect match. In our case, we apply it to the segmented LV
of our method and the ground truth LV so as to figure out how close we are to the correct result.

\figurename ~\ref{dice_g} presents this metric applied to each slice of the considered patient in the dataset with respect to each experiment's setting. Additionally, we present the dice of each model: MF, RF, and the union both so as to see closely how each model performs.  There are 9 slices where the first one is used to train the algorithm which then infers subsequent slices. So graphs in these figures are the dice score from slice 2 to 9.

\subsubsection{Patient 11}
In the first experiment's graph (\figurename ~\ref{dice_g}(a)), dice scores for all models globally decrease as we move through slice except the Random Forest that gets up around middle slices but falls afterwards. This is consistent with the fact that distant slices less similar. But surprisingly dice scores are quite good for learning only from slice 1. This not only confirms our intuition but goes farther to the point that learning only from the first slice can give us good result. Dices score of overall slices are respectively: 0.869, 0.865 and 0.871 for the MF, RF and the combination of  both. That's a very good starting point. We notice that the combination of  both models improves the dice score by a small fraction. 

The second experiment's graph (\figurename ~\ref{dice_g}(b)) where we apply the post-processing step shows a great improvement in the dice score which now fluctuates between high values. Dices score of overall slices are now respectively: 0.947, 0.920 and 0.948 for the MF, RF and the combination of  both. This is a huge jump. Here we notice that the combination of models performs slightly better than the MF one.

Concerning the third experiment's graph (\figurename ~\ref{dice_g}(c)), the RF result is close to that of MF. This normal as we are only updating the RF. A question that might raised is why have we only experimented the RF update only ? The answer is that updating the MF results in poor accuracy.

\subsubsection{The sample of 45 patients}
\tablename~\ref{res_tab} reports the mean $\pm$ variance of dice scores obtained after running the three experiments on a sample of 45 patients. We still notice a significant jump in accuracy after post-processing.
But the update don't show a significative improvement.

\section{Related work}

The studied problem comes from a challenge where a total of 10 teams competed.
The leaderboard \footnote{$https://www.creatis.insa-lyon.fr/Challenge/acdc/miccai_results.html$}
has better results than ours. But our approach is less data-hungry than theirs because we used approximately
10\% of labels if we consider each MRI producing around 10 slices.This is very cost effective while competitors used CNN models that use all labels.

Some works use the same approach as ours like \cite{slic-seg} where they developed a system for placenta segmentation from MR images using a single slice label with scribbles.
We can't compared their results with ours as images are different and they didn't provide a way to get their dataset. However they were able to get state-of-the-art results for the placenta segmentation from fetal MRI.
This shows that this strategy is very promising.

A drawback of our method is that it fails in case of misalignment of slices. Because features are mainly pixel coordinates. One solution could be to use other features. In our studies, we have in fact used features from Local Binary Pattern (LBP)\cite{lbp} and Histogram of Gradients (HOG)\cite{hog}. However, results were very poor compared to the current one. Perhaps because, a single slice label is not enough for these features to expose significant structure to the used forest algorithms.

\section{Conclusion and Perspectives}
We presented a method for segmenting the left ventricle in cardiac cine-MR images while reducing manual label effort. This segmentation is in practice used to detect some heart failures.
This approach leverages similarities and alignment between slices to initiate the system by training it on a single manually labeled slice. Then the system follows an infer-and-learn scheme across all slices without additional manual label. We experiment this method using Mondrian and Random Forests and their combination. Results confirm that idea and show more. Indeed, after learning the first slice, the MF gives its best results when inferring subsequent slices without updates while its counterpart shows slighty better results with updates. 

Recently, a new paradigm has been developed, namely data programming \cite{dataprog} which aims in reducing manual label effort. It has three steps:
\begin{enumerate}
  \item Building a training set by writing function to generate labels.
  \item Modeling this training set to denoise it.
  \item Training a noise-aware discriminative model.
\end{enumerate}

They found that in some cases, the discriminative model trained on the denoised training set performs better than it is trained directly with true labels. So our next research direction is to use this data-programming paradigm considering our approach as a labeling functions and results as noisy labels.


%
%
%



%

\end{document}